\pdfoutput=1

\documentclass[11pt]{article}

\usepackage{algorithm}
\usepackage[preprint]{acl}
\usepackage{enumitem}
\usepackage{algorithmicx}
\usepackage{times}
\usepackage{latexsym}
\usepackage{booktabs}
\usepackage[T1]{fontenc}

\usepackage[utf8]{inputenc}

\usepackage{microtype}

\usepackage{inconsolata}

\usepackage{graphicx}
\usepackage{amsmath}
\usepackage{amssymb}
\title{Meta-Learning Reinforcement Learning for Crypto-Return Prediction}

\author{
 \textbf{Junqiao Wang\textsuperscript{* 1}},
 \textbf{Zhaoyang Guan\textsuperscript{* 2}},
 \textbf{Chuhan Cheng\textsuperscript{* 1}},
 \textbf{Guanyu Liu\textsuperscript{9}},\\
 \textbf{Tianze Xia\textsuperscript{3}},
 \textbf{Xianzhi Li\textsuperscript{4}},
 \textbf{Shuo Yin\textsuperscript{7}},\\
\textbf{Xinyuan Song\textsuperscript{8}},
 \textbf{Tianyu Shi\textsuperscript{5}},
 \textbf{Alex Lee\textsuperscript{6}},
\\
\\
 \textsuperscript{1}Sichuan University,
 \textsuperscript{2}Shanghai Artificial Intelligence Laboratory,
 \textsuperscript{3}Northwestern University, \\
 \textsuperscript{4}Huazhong University of Science and Technology,
 \textsuperscript{5}Queen's University, \\
 \textsuperscript{6}University of Toronto,
 \textsuperscript{7}TrueNorth,
 \textsuperscript{8}Tsinghua University, \\
 \textsuperscript{9}Emory University,
 \textsuperscript{10}University of Macau
\\
 \small{
   \textbf{Correspondence:} \href{ty.shi@mail.utoronto.ca}{ty.shi@mail.utoronto.ca}
 }
}
\vspace{+4pt}
\begin{document}
\maketitle

\begin{abstract}
Predicting cryptocurrency returns is notoriously difficult: price movements are driven by a fast-shifting blend of on-chain activity, news flow, and social sentiment, while labeled training data are scarce and expensive. In this paper, we present \textbf{Meta-RL-Crypto}, a unified transformer-based architecture that unifies meta-learning and reinforcement learning (RL) to create a fully self-improving trading agent. Starting from a vanilla instruction-tuned LLM, the agent iteratively alternates between three roles—actor, judge, and meta-judge—in a closed-loop architecture. This learning process requires no additional human supervision. It can leverage multimodal market inputs and internal preference feedback. The agent in the system continuously refines both the trading policy and evaluation criteria. Experiments across diverse market regimes demonstrate that Meta-RL-Crypto shows good performance on the technical indicators of the real market and outperforming other LLM-based baselines.
\vspace{\baselineskip} 
\end{abstract}

\section{Introduction}
Large Language Models (LLMs) have shown promise in financial tasks like sentiment analysis and time-series reasoning\cite{makri2025ethereum}, but face two key challenges: (i) reducing dependence on human-curated data, and (ii) reliably forecasting volatile cryptocurrency markets influenced by on-chain data, news, and social sentiment.

\begin{figure*}[htbp]
    \centering
    \includegraphics[width=0.75\linewidth]{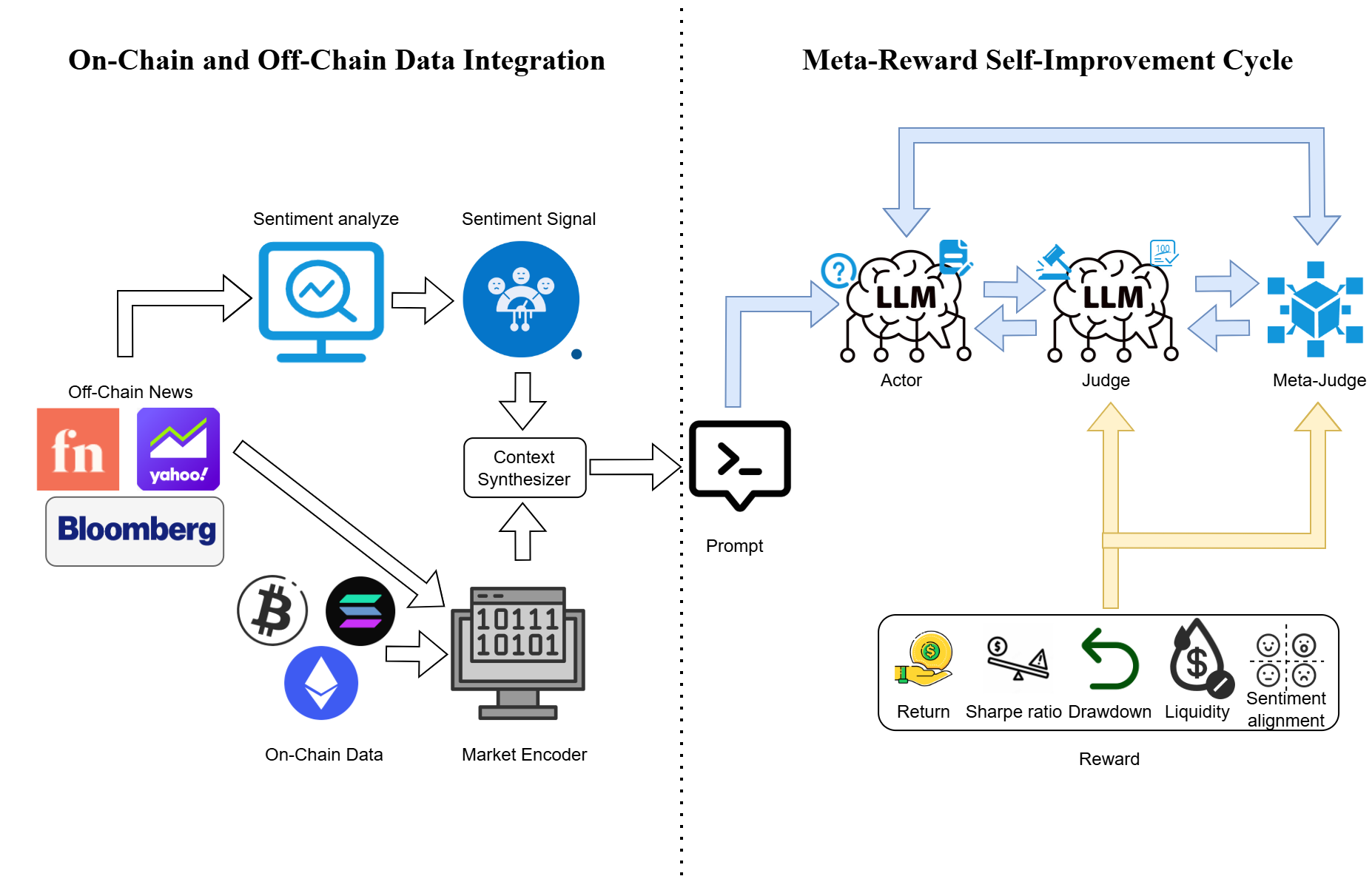}
    \caption{ \textbf{Overall Architecture of Meta-RL-Crypto.}
            The system consists of a shared LLM that cyclically adopts the roles of \textit{Actor}, \textit{Judge}, and \textit{Meta-Judge}.
            Market signals (on-chain metrics, off-chain news, sentiment scores) are encoded into structured prompts used by the Actor to generate forecasts.
            Each prediction is then scored by the Judge using a multi-dimensional reward vector, which the Meta-Judge uses to enforce preference consistency and evaluate the Judge itself.}
    \label{fig:arch}
\end{figure*}

To address the challenges in cryptocurrency prediction, we propose Meta-RL-Crypto, a framework that combines meta-reward-driven self-improvement with multi-modal trading intelligence in Figure \ref{fig:arch}. The central innovation is a triple-loop learning process in which a single large language model (LLM) takes on three distinct roles. First, the Actor processes on-chain metrics (such as gas fees and transaction graphs), news, and sentiment to generate next-day forecasts for crypto-assets \cite{meister2024gasfees, liu2025llm4fts}. Next, the Judge evaluates these forecasts using a multi-objective reward vector, which incorporates absolute returns, the Sharpe ratio, drawdown control, and sentiment alignment. This approach reduces potential bias from relying on a single metric. Finally, the Meta-Judge refines the Judge's reward policy through preference comparisons, helping to prevent reward drift and length bias \cite{lee2024self_improving, wu2024meta_rewarding}. This closed-loop system enables continuous self-improvement without human intervention, allowing it to adapt dynamically to shifts in the market.

Our contributions are as follows:

\begin{itemize}[left = 0em]
\item \textbf{Unified Meta-Learning RL Framework:} We combine meta-reward self-improvement with crypto-specific trading goals, creating a single actor–judge–meta-judge loop that works on raw multimodal data.

\item \textbf{Multi-Objective Reward Design:} A supervisory system using financial and emotional incentives helps prevent reward hacking and promotes better trading behavior.

\item \textbf{Empirical Validation:} Initial experiments with BTC, ETH, and SOL daily trading show that Meta-RL-Crypto outperforms models like Informer and PatchTST, as well as traditional indicators like MACD, while matching or surpassing GPT-4-based baselines—without using extra human-labeled data.
\end{itemize}

\section{Related Works}

\subsection{LLMs for Financial Analysis and Cryptocurrency Markets}
Recent advancements in large language models (LLMs) have transformed financial-market research, though most studies have concentrated on equities, leaving the rich on-chain data of cryptocurrencies less explored~\cite{roumeliotis2024llm_sentiment_crypto, li-etal-2024-cryptotrade}. Early time-series forecasting in finance evolved from traditional econometrics to machine learning and reinforcement learning methods, with Long Short-Term Memory networks proving effective for sequential price data~\cite{fjellstrom2022lstm_ensemble_finance, siami2018arima_vs_lstm, sezer2019dl_financial_timeseries}. Transformer-based models, such as Informer~\cite{zhou2021informer}, AutoFormer~\cite{wu2021autoformer}, PatchTST~\cite{nie2022patchtst}, and TimesNet~\cite{liu2023timesnet}, now lead the field in long-horizon forecasting. In addition, domain-specific LLMs—such as FinGPT~\cite{liu2023fingpt}, BloombergGPT~\cite{yang2023bloomberggpt}, and FinMA~\cite{xie2023pixiu}—expand LLM capabilities to sentiment analysis, entity recognition, question answering, and market prediction. While these techniques show promise for more robust decision-making, their application to cryptocurrency markets remains underexplored. Our work addresses this gap by integrating both on-chain and off-chain signals with self-reflective LLM agents to navigate the volatile, information-rich landscape of cryptocurrency assets.

Reinforcement learning has shown promise in addressing the challenges of delayed feedback, making it well-suited for tasks like comment ranking, where the effects of actions may not be apparent until later. For instance, Yahoo’s ranking system used contextual bandits to dynamically optimize comment visibility based on user interactions \cite{kulkarni2020context}. In finance, RL agents have used social media sentiment signals, such as those from Twitter (X), to inform trading decisions and adjust portfolios \cite{xiao2018trading}. Building on these insights, our work treats comment ranking as an RL problem by modeling tweet-comment pairs as sequential episodes and using subsequent market returns as delayed reward signals. This setup facilitates long-term credit assignment, aligning model decisions with financial outcomes and providing a more meaningful evaluation of comment informativeness. Reward modeling in financial text analysis has traditionally focused on price prediction tasks \cite{jiang2017deep}, where models predict asset price movements based on textual data. Recent approaches have aimed to align textual signals with economic indicators or other quantitative metrics \cite{yang2020modelling}, but still depend on extensive manual supervision and predefined labels. 

\subsection{Meta-Learning in Reinforcement Learning}
Recent approaches have explored how a single instruction-tuned seed model can generate its own training signals through self-play~\cite{chen2024spin, fang2025serl, zhang2024rest_mcts, shinn2024crescent}. In these frameworks, the model assumes three roles: the Actor generates multiple candidate responses for each user prompt; the Judge evaluates these candidates using a rubric and assigns a score; and the Meta-Judge compares the judgments produced by the Judge, selecting the best one to create preference data for training the Judge. This cyclical process enhances the model's ability to refine its performance over time.


\section{Multi-Reward CryptoTrade Framework}
\label{sec:multireward}

This section shows the multi-reward architecture of Meta-Rewarding learning loop.  We extract several orthogonal reward channels from a unified data pipeline. These channel-specific rewards are combined by a meta-judge to provide stable, fine-grained feedback for training the actor.

\subsection{Data Collection}
\label{ssec:data}

We retain the dual on-/off-chain data strategy of the original system:

\paragraph{On-chain channel.}%
We gather macro and micro blockchain indicators from two complementary public APIs:
\begin{itemize}[left = 0em]
    \item \textbf{CoinMarketCap~\cite{coinmarketcap_api} (CMC).}  
          CMC’s REST v2 endpoint provides daily OHLC price bars, traded volume, and fully-diluted market capitalization for BTC, ETH, and SOL.  
          These series capture long-horizon valuation cycles and liquidity regimes.
    \item \textbf{Dune Analytics~\cite{dune_analytics} (GraphQL).}  
          We execute SQL-backed dashboards to extract network-level activity:  
          (i) total transaction count,  
          (ii) unique active wallets,  
          (iii) aggregate value transferred (USD),  
          (iv) mean \& median gas price (Gwei), and  
          (v) total gas consumed.  
          These metrics illuminate congestion patterns, fee pressure, and real-time liquidity that directly influence execution costs and slippage.
\end{itemize}

\paragraph{Off-chain channel.}%

We utilize the GNews API~\cite{gnews_api} to harvest a daily corpus of news reports for each target cryptocurrency.
Because GNews federates headlines from the Google News index, this single endpoint provides wide coverage while preserving source metadata (publisher, timestamp, URL).
To maximise informational reliability, we post-filter the feed to retain only articles published by high-credibility financial outlets—Bloomberg, Yahoo Finance, Reuters, crypto.news, and comparable tier-one crypto media~\cite{bloomberg, yahoofinance, reuters, cryptonews}.
Each retained item is stored with its full headline and body text, and then de-duplicated via 64-bit SimHash~\cite{charikar2002simhash} to eliminate near-identical wire-service repeats.
The resulting cleaned stream offers a time-stamped snapshot of prevailing market discourse; subsequent sentiment and relevance scoring 
allows the model to anticipate price moves that frequently follow shifts in collective narrative.

\subsection{Reward Channel Construction}
\label{ssec:rewards}

From the fused data stream, we compute the following daily reward signals, normalized to the range $[-1,1]$, and reported after the close of trade $t$. The \textbf{Return-Based Reward $\mathrm{R}_{\mathrm{return}}$} is the realized net percentage gain after fees, assuming the actor’s long/short allocation $\alpha_t$ is executed. The \textbf{Risk-Adjusted Reward $\mathrm{R}_{\mathrm{sharpe}}$} represents the incremental Sharpe ratio contribution of the position, estimated using an exponentially-weighted variance window. The \textbf{Drawdown Reward $\mathrm{R}_{\mathrm{dd}}$} is a penalty proportional to the maximum intra-day drawdown induced by $\alpha_t$. The \textbf{Liquidity Reward $\mathrm{R}_{\mathrm{liq}}$} provides a bonus for selecting position sizes that keep expected slippage below a threshold, derived from on-chain volume and gas costs. Lastly, the \textbf{Sentiment Alignment Reward $\mathrm{R}_{\mathrm{sent}}$} is the cosine similarity between the actor’s textual rationale and a sentiment vector extracted from aggregated news using a frozen sentiment-LM~\cite{wang2020sentilm}.

Each channel targets a distinct desideratum—profitability, risk control, market impact, and information utilization—mitigating single-metric reward hacking.

\begin{figure*}[ht]
    \centering
    \includegraphics[width=1.00\textwidth]{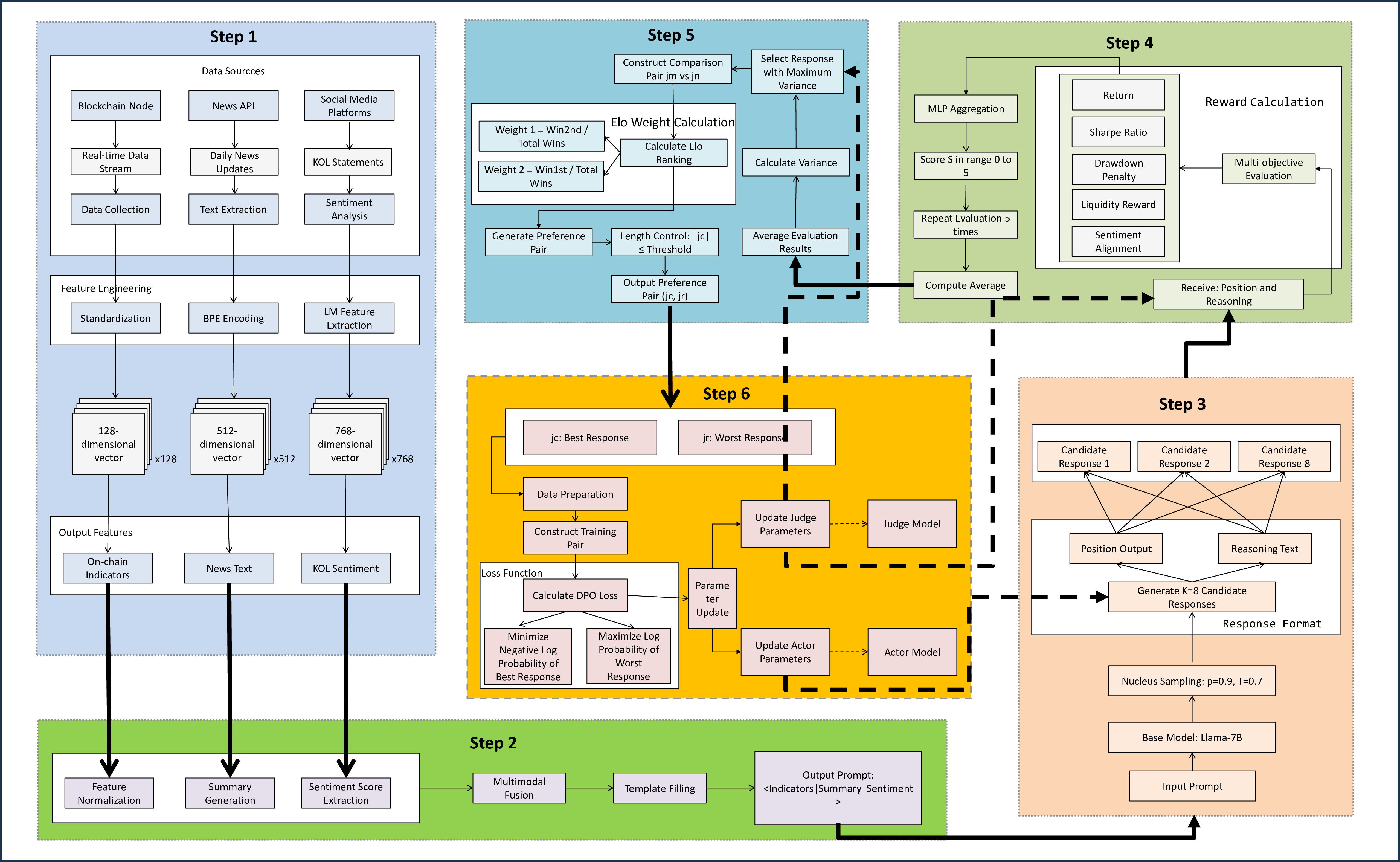}
    \caption{
    \textbf{Meta-RL-Crypto Architecture.} The diagram illustrates the cyclical roles of Actor, Judge, and Meta-Judge, showing how data is processed through the system. 
    Each role contributes to improving the model’s performance, from generating forecasts (Actor) to evaluating them (Judge), and refining evaluations (Meta-Judge).
    }
    \label{fig:workflow}
\end{figure*}

\begin{table*}
\caption{
\textbf{Market Regime Test Periods for BTC, ETH, and SOL.}
The table shows the three market conditions (bearish, sideways, bullish) used for evaluating Meta-RL-Crypto, with opening and closing prices demonstrating each regime's characteristics. Percentage changes reflect the overall trend direction and magnitude during each test period.
}
\centering
\label{tab:1}
\begin{tabular}{llccccr} 
    \toprule 
    \textbf{Type} & \textbf{Split} & \textbf{Start} & \textbf{End} & \textbf{Open} & \textbf{Close} & \textbf{Trend} \\
    \midrule 
    BTC & Test Bearish & 2025-04-08 & 2025-05-23 & 79,163.24 & 107,318.30 & \textcolor{red}{+35.56\%} \\
    & Test Sideways & 2025-03-10 & 2025-04-06 & 80,734.48 & 78,430.00 & \textcolor{green}{-2.85\%} \\
    & Test Bullish & 2025-01-30 & 2025-02-28 & 103,733.25 & 84,349.94 & \textcolor{green}{-18.68\%} \\
    \addlinespace 
    ETH & Test Bearish & 2025-01-07 & 2025-03-11 & 3,687.44 & 1,923.43 & \textcolor{green}{-47.82\%} \\
    & Test Sideways & 2025-05-29 & 2025-06-11 & 2,681.61 & 2,771.61 & \textcolor{red}{+3.36\%} \\
    & Test Bullish & 2025-04-08 & 2025-05-07 & 1,553.04 & 1,811.11 & \textcolor{red}{+16.62\%} \\
    \addlinespace
    SOL & Test Bearish & 2025-01-26 & 2025-02-27 & 256.40 & 137.68 & \textcolor{green}{-46.29\%} \\
    & Test Sideways & 2025-03-11 & 2025-04-06 & 118.32 & 105.91 & \textcolor{green}{-10.49\%} \\
    & Test Bullish & 2025-04-08 & 2025-05-14 & 106.98 & 176.64 & \textcolor{red}{+65.12\%} \\
    \bottomrule 
\end{tabular}
\end{table*}

\subsection{Reward Aggregation}
\label{ssec:reward_agg}

\noindent\textbf{Actor preference dataset with length control.}
Given the prompt:
\begin{equation}
\begin{aligned}
x_t&=\langle\mathrm{on-chain metrics},\\
&\mathrm{news digest},\mathrm{sentiment snapshot}\rangle,
\end{aligned}
\end{equation}
the actor model generates $K$ candidates $\{y_t^{(k)}\}_{k=1}^K$ using nucleus sampling ($p=0.9$, $T=0.7$). Each candidate is evaluated $N$ times, with malformed outputs discarded and scores averaged to $\bar{S}_{t,k}$. 

Define $S_{\max}$ and $S_{\min}$ as the highest/lowest scores. A tunable threshold $\rho\in[0,1]$ partitions scores:
\begin{equation}
\begin{aligned}
\mathrm{Top-tier}&=[(1-\rho)S_{\max}+\rho S_{\min},S_{\max}],\\
\mathrm{Low-tier}&=[S_{\min},(1-\rho)S_{\min}+\rho S_{\max}].
\end{aligned}
\end{equation}
Select the shortest top-tier candidate as positive $y_c$ and longest low-tier as negative $y_r$.

\noindent\textbf{Judge preference dataset via Elo aggregation.}
Adapt Elo rating with: Dynamic $K_t=K_{\mathrm{base}}\cdot(1+\sigma_t/\sigma_{\max})$ and Non-zero-sum adjustment as
    \begin{equation}
    \begin{aligned}
    \Delta\mathrm{Elo}&=K\cdot\Big(I[r_{\mathrm{model}}>r_{\mathrm{market}}]\\
    &-\frac{1}{1+10^{(S_{\mathrm{opponent}}-S_{\mathrm{model}})/400}}\Big)
    \end{aligned}
    \end{equation}

For high-variance forecasts, form judgment pairs $(j^{(m)},j^{(n)})$ and record wins in matrix $B_{mn}$ with order-normalized weights $\omega_1,\omega_2$. Compute Elo scores by maximizing:
\begin{equation}
\sum_{m,n}B_{mn}\log\sigma(\varepsilon_m-\varepsilon_n).
\end{equation}
Select highest- and lowest-scoring judgments as positive/negative samples, discarding verbose pairs.

\subsection{Generalized Preference-based Reinforcement Optimization}

We incorporate a reinforcement learning strategy inspired by Generalized Preference-based Reinforcement Optimization (GPRO)~\cite{pmlr-v235-tang24b} to optimize the actor, judge, and meta-judge in a closed feedback loop. At each training step, the actor samples two candidate responses, which are evaluated based on precomputed criteria. These evaluations result in reward vectors $(\mathbf{r}_t^{(1)}, \mathbf{r}_t^{(2)})$, which are passed through an aggregation MLP to produce scalar rewards $\hat{r}_t^{(i)} = f_{\mathrm{agg}}(\mathbf{r}_t^{(i)})$. These scalar rewards form input pairs $(r_t^{(1)}, r_t^{(2)})$ for the meta-judge, which models the preference between them. The meta-judge is trained with a DPO-style loss:

\[
\mathcal{L}_{\mathrm{meta}} = -\log p,
\]

where $p = \sigma \left( M_\phi(r_t^{(1)}, r_t^{(2)}) \right)$ and $\sigma(\cdot)$ is the sigmoid function. The preference probability $p$ should approach 1 when $r_t^{(1)}$ is truly better, guiding the meta-judge to favor higher-quality outputs. The meta-judge's parameters $\phi$ are updated accordingly.

To efficiently propagate the preference signal, a lightweight judge $M_\theta$ is trained to approximate the meta-judge’s behavior. The judge is trained using a supervised regression objective to minimize the discrepancy between the meta-judge and judge’s preferences:

\[
\mathcal{L}_{\mathrm{align}} = \mathbb{E} \left[ \left( M_\phi(r_t^{(1)}, r_t^{(2)}) - M_\theta(r_t^{(1)}, r_t^{(2)}) \right)^2 \right].
\]

This allows the judge to distill the preference knowledge from the meta-judge, enabling efficient evaluation during actor fine-tuning.

For actor optimization, preference pairs $(j_c, j_r)$ are constructed from the actor’s outputs, where $j_c$ is the candidate with the higher Elo aggregation score, and $j_r$ has the lower score. These pairs are used to optimize the actor’s policy, encouraging it to produce outputs similar to $j_c$ while suppressing those resembling $j_r$. The training objective is:

\[
\mathcal{L}_{\mathrm{actor}} = \mathbb{E}\left[\log \frac{\exp(\pi_\theta(j_c)/\beta)}{\exp(\pi_\theta(j_c)/\beta) + \exp(\pi_\theta(j_r)/\beta)}\right],
\]

where $\pi_\theta$ is the actor policy and $\beta$ is a temperature parameter.

Our approach mitigates overfitting through two mechanisms: (1) the self-supervised preference signals are comparative and robust to noise, and (2) joint training of generation and evaluation modules regularizes each other. As shown in our experiments, the model exhibits strong generalization despite frequent market regime shifts.

\begin{table*}[!ht]
\centering
\caption{
\textbf{Comparative Performance Against State-of-the-art Baselines.}
Our Meta-RL approach demonstrates superior risk-adjusted returns compared to state-of-the-art language models (GPT-4, Gemini) and specialized financial AI systems (DeepSeek, DMind), with particularly strong performance in challenging bear market conditions (-8\% vs. -12\% to -22\%).
}
\label{tab:3}
\resizebox{\textwidth}{!}{
\begin{tabular}{lccccccccc}
\toprule
\textbf{Model} & \multicolumn{3}{c}{\textbf{Total Return (\%)}} & \multicolumn{3}{c}{\textbf{Daily Return (\%)}} & \multicolumn{3}{c}{\textbf{Sharpe Ratio}} \\

\cmidrule(lr){2-4} \cmidrule(lr){5-7} \cmidrule(lr){8-10}
& \textbf{Bull} & \textbf{Sideways} & \textbf{Bear} & \textbf{Bull} & \textbf{Sideways} & \textbf{Bear} & \textbf{Bull} & \textbf{Sideways} & \textbf{Bear} \\
\midrule
DMind & 28.00 & -3.20 & -20.50 & 0.38 & -0.04 & -0.28 & 0.18 & -0.06 & -0.18 \\
Gemini & 32.00 & 1.80 & -15.00 & 0.42 & 0.02 & -0.20 & 0.22 & 0.01 & -0.12 \\
ChatGPT-4 & 25.00 & -5.00 & -22.00 & 0.35 & -0.07 & -0.30 & 0.15 & -0.10 & -0.20 \\
DeepSeek & 35.00 & 0.50 & -12.00 & 0.45 & 0.01 & -0.16 & 0.25 & 0.00 & -0.10 \\
\midrule
\textbf{Meta-RL-Crypto (Ours)} & \textbf{42.00} & \textbf{4.50} & \textbf{-8.00} & \textbf{0.52} & \textbf{0.06} & \textbf{-0.10} & \textbf{0.30} & \textbf{0.08} & \textbf{-0.05} \\
\bottomrule
\end{tabular}}
\end{table*}

\begin{table*}[!ht]
\centering
\caption{Market Interpretability Evaluation Scores (0-1 scale)}
\label{tab:interpretability}
\begin{tabular}{lcccc}
\toprule
\textbf{Metric} & \textbf{MACD} & \textbf{LSTM} & \textbf{GPT-4} & \textbf{Meta-RL-Crypto (Ours)} \\
\midrule
Market Relevance & 0.42 ± 0.11 & 0.38 ± 0.09 & 0.67 ± 0.13 & \textbf{0.82 ± 0.07} \\
Risk-Awareness & 0.51 ± 0.12 & 0.45 ± 0.10 & 0.59 ± 0.15 & \textbf{0.85 ± 0.06} \\
Adaptive Rationale & 0.18 ± 0.07 & 0.31 ± 0.08 & 0.63 ± 0.14 & \textbf{0.88 ± 0.05} \\
\bottomrule
\end{tabular}
\end{table*}

\section{Experiments and analysis}
\label{sec:exp}
\subsection{Experiment Settings}


We fine-tune a Llama-7B~\cite{touvron2023llama} model using a Meta-Reward Reinforcement Learning (Meta-RL) framework. All data is segmented by market regimes: bullish, bearish, and sideways. We use real 2025 price trajectories without simulation. To avoid look-ahead bias~\cite{bailey2016probability}, the model only sees historical data at each prediction step. The backtest simulation starts with a \$1,000,000 portfolio, comprising 50\% cash reserve (\$500k) and equal initial allocations to BTC, ETH, and SOL (\$166.7k each). Thereafter, daily rebalancing is fully governed by the actor’s normalized position signal $\alpha_t \in [-1,1]$ (generated by $y_t$): positive values prompt proportional buying across assets using available cash, while negative values trigger proportional reduction of existing holdings. Transactions incur a 10 basis point fee and asset-specific slippage, modeled as $\mathcal{N}(0, 0.05\%)$ for BTC/ETH and $\mathcal{N}(0, 0.12\%)$ for SOL, based on historical order book data.  To rigorously evaluate Meta-RL-Crypto’s adaptability across diverse market conditions, we construct test periods for Bitcoin (BTC), Ethereum (ETH), and Solana (SOL) spanning three distinct regimes: bearish, sideways, and bullish. As shown in Table~\ref{tab:1}, each regime is characterized by its start/end dates, opening/closing prices, and trend magnitude (percentage change).

We report four standard metrics: 
\begin{itemize}
    \item \textbf{Total Return}: $R = \frac{w_{\mathrm{end}} - w_{\mathrm{start}}}{w_{\mathrm{start}}}$, where $w_{\mathrm{start}}$ and $w_{\mathrm{end}}$ denote the initial and final wealth, respectively. 
    \item \textbf{Sharpe Ratio}: $S = \frac{\bar{r}}{\sigma}$
calculated using daily log-returns ($r_t = \ln(w_t/w_{t-1})$), where $\bar{r}$ is the sample mean, $\sigma$ is the sample standard deviation, and the risk-free rate $r_f = 0$. 
    \item \textbf{Daily Return Mean: } $\bar{r} = \frac{1}{T}\sum_{t=1}^T r_t$, where $T$ is the total number of days, and $r_t$ is the log-return on day $t$.
\end{itemize}


\subsection{Results}

The results of our experimental evaluation are presented in two key tables. \autoref{tab:3} compares the performance of our Meta-RL-Crypto model with several state-of-the-art baselines, including GPT-4, Gemini, DeepSeek, and DMind. Our approach outperforms these models, especially in challenging bear market conditions, achieving a \emph{total return} of $-8\%$ compared to $-12\%$ to $-22\%$ for the baselines. Furthermore, Meta-RL-Crypto shows superior \emph{risk-adjusted returns} with a \emph{Sharpe ratio} of $0.30$, surpassing other models across different market conditions.

In \autoref{tab:interpretability}, we evaluate the \emph{market interpretability} of different models on a $0$-$1$ scale. Meta-RL-Crypto stands out by achieving the highest scores across three important metrics: \emph{Market Relevance}, \emph{Risk-Awareness}, and \emph{Adaptive Rationale}, with values of $0.82 \pm 0.07$, $0.85 \pm 0.06$, and $0.88 \pm 0.05$, respectively. These results highlight the model's ability to interpret market conditions effectively and adapt its rationale in real-time, offering significant improvements over traditional models like MACD, LSTM, and even GPT-4.

Overall, these experimental results demonstrate that our Meta-RL-Crypto model not only provides superior trading performance but also excels in terms of market interpretability and adaptability, setting a new benchmark for financial AI systems.



\section{Conclusion}
\label{ssec:conclusion}
In this paper, we present Meta-RL-Crypto, a meta-learning reinforcement learning framework designed for cryptocurrency return prediction. Meta-RL-Crypto operates through a self-improving, closed-loop architecture, refining its trading policy without requiring human annotations or external supervision.  Our approach combines multi-modal data, integrating both on-chain blockchain metrics and off-chain sentiment signals to better capture the volatility and complexity of cryptocurrency markets. Additionally, we introduce a multi-objective reward design, which includes profitability, risk control, liquidity, and sentiment alignment. This integrated reward structure prevents reward hacking and improves overall trading behavior. Overall, Meta-RL-Crypto advances AI applications in cryptocurrency prediction by providing a robust, self-improving system that adapts to dynamic market conditions. We believe this framework can improve trading strategies in fast-changing financial environments.

\clearpage
\appendix

\section{Human Expert Evaluation}
\label{app:expert_eval}

To complement automatic metrics, we conducted a structured human evaluation on the reasoning quality of model outputs using domain-expert judgment. This appendix details the expert sources, evaluation criteria, rating process, and analysis methods.

\subsection{Expert Panel Composition}
We recruited five expert annotators with diverse yet relevant backgrounds: two professional crypto analysts with over three years of industry experience, two PhD students specializing in quantitative finance and machine learning, and one quantitative researcher at a hedge fund. All evaluators were blind to the model identities during scoring to prevent bias.

\subsection{Evaluation Criteria}
Each model output was evaluated across four dimensions using a 5-point Likert scale (1 = poor, 5 = excellent):

\begin{table}[H]
\centering
\caption{Expert Scoring Criteria}
\label{tab:expert_criteria}
\resizebox{\columnwidth}{!}{
\begin{tabular}{ll}
\toprule
\textbf{Dimension} & \textbf{Description} \\
\midrule
Soundness & Logical validity and internal coherence of the reasoning \\
Consistency & Agreement with stated assumptions or prior context \\
Completeness & Coverage of necessary reasoning steps or evidence \\
Relevance & Pertinence of the reasoning to the prompt question \\
\bottomrule
\end{tabular}}
\end{table}

\subsection{Scoring Examples}
To ensure transparency, Table~\ref{tab:scoring_examples} illustrates typical scoring decisions made by experts on selected model outputs.

\begin{table}[H]
\centering
\caption{Sample Model Outputs and Expert Scores. \textbf{Note:} S = Soundness, C = Consistency, Cp = Completeness, R = Relevance}
\label{tab:scoring_examples}
\resizebox{\columnwidth}{!}{
\begin{tabular}{p{9cm}cccc}
\toprule
\textbf{Model Output (Excerpt)} & \textbf{S} & \textbf{C} & \textbf{Cp} & \textbf{R} \\
\midrule
"Given BTC dominance rose while ETH volume fell, we expect rotation into altcoins..." & 5 & 5 & 4 & 5 \\
"The token price will increase because it's Monday and usually prices go up..." & 2 & 2 & 1 & 2 \\
\bottomrule
\end{tabular}}

\end{table}

\subsection{Controlling for Subjectivity}
To reduce individual bias and assess inter-rater reliability, we report both average scores and agreement metrics. We compute the Kendall’s W coefficient~\cite{kendall1948rank} and Krippendorff’s alpha~\cite{krippendorff2004reliability} to measure rating consistency across experts. For our evaluation set, we obtain Kendall’s $W = 0.78$ and Krippendorff’s $\alpha = 0.71$, indicating substantial agreement.

\subsection{Comparative Results}
Table~\ref{tab:expert_eval} presents the aggregated expert scores for each model over 50 randomly selected prompts. Meta-RL-Crypto consistently outperforms both ChatGPT-3.5 and GPT-4 across all dimensions, particularly in soundness and completeness. The high soundness score reinforces the reliability of the model's reasoning process and highlights its advantage in structured, logic-driven inference.

\begin{table}[H]
\centering
\caption{Expert Evaluation on Reasoning Quality (Avg. over 50 prompts)}
\label{tab:expert_eval}
\resizebox{\columnwidth}{!}{
\begin{tabular}{lcccc}
\toprule
\textbf{Model} & \textbf{Soundness} & \textbf{Completeness} & \textbf{Relevance} & \textbf{Avg. Score} \\
\midrule
\textbf{Meta-RL-Crypto (Ours)} & 4.6 & 4.4 & 4.5 & \textcolor{green}{\textbf{4.5}} \\
\textbf{ChatGPT-3.5}    & 3.9 & 3.5 & 4.0 & 3.8 \\
\textbf{GPT-4} (zero-shot) & 4.2 & 4.0 & 4.1 & 4.1 \\
\bottomrule
\end{tabular}}
\end{table}

\end{document}